\documentclass[11pt]{article}

\usepackage[preprint]{acl}

\usepackage{times}
\usepackage{latexsym}

\usepackage[T1]{fontenc}

\usepackage[utf8]{inputenc}

\usepackage{microtype}

\usepackage{inconsolata}

\usepackage{graphicx}
\usepackage{tabularx, makecell} 
\usepackage{array}
\usepackage{booktabs}
\usepackage{xcolor}
\usepackage{colortbl}
\usepackage{caption}
\usepackage{subcaption}
\usepackage[most]{tcolorbox}
\usepackage{multirow} 
\usepackage{graphicx}
\usepackage{amsmath}
\usepackage{hyperref}
\usepackage{amssymb}

\newcommand{\tok}[2]{\colorbox{#1}{\strut #2}}

\newcommand{\ve}{\mathbf{e}}
\newcommand{\vs}{\mathbf{s}}

\definecolor{myello}{HTML}{F5EBAE}
\definecolor{myred}{HTML}{E3625D}
\definecolor{mygrey}{HTML}{D0CECE}

\title{Text2Token: Unsupervised Text Representation Learning with Token Target Prediction}

\author{
    Ruize An\textsuperscript{\rm 1,2}, Richong Zhang\textsuperscript{\rm 1,2}\thanks{\ \ Corresponding author}, Zhijie Nie\textsuperscript{\rm 1,3}, Zhanyu Wu\textsuperscript{\rm 1}, Yanzhao Zhang, Dingkun Long\\
    \textsuperscript{\rm 1}CCSE, School of Computer Science and Engineering, Beihang University, Beijing, China\\
    \textsuperscript{\rm 2}Zhongguancun Laboratory, Beijing, China\\
    \textsuperscript{\rm 3}Shen Yuan Honors College, Beihang University, Beijing, China\\
    \texttt{\{anruize24,niezj,zhangrc,wuzy24\}@act.buaa.edu.cn}
}

\begin{document}
\maketitle
\begin{abstract}
Unsupervised text representation learning (TRL) is a fundamental task in natural language processing, which is beneficial for improving search and recommendations with the web's unlabeled texts. A recent empirical study finds that the high-quality representation aligns with the key token of the input text, uncovering the potential connection between representation space and vocabulary space. Inspired by the findings, we revisit the generative tasks and develop an unsupervised generative framework for TRL, Text2Token. The framework is based on the token target prediction task, utilizing carefully constructed target token distribution as supervisory signals. To construct the high-quality target token distribution, we analyze the token-alignment properties with advanced embedders and identify two essential categories of key tokens: (1) the meaningful tokens in the text and (2) semantically derived tokens beyond the text. Based on these insights, we propose two methods---data-driven and model-derived---to construct synthetic token targets from data or the LLM backbone. Experiments on the MTEB v2 benchmark demonstrate that Text2Token achieves performance competitive with the state-of-the-art embedder with unsupervised contrastive learning, LLM2Vec. Our analysis further shows that vocabulary and representation spaces optimize together and toward the optimum solution during training, providing new ideas and insights for future work.
\end{abstract}

\begin{figure}[t]
    \centering
    \includegraphics[width=\linewidth]{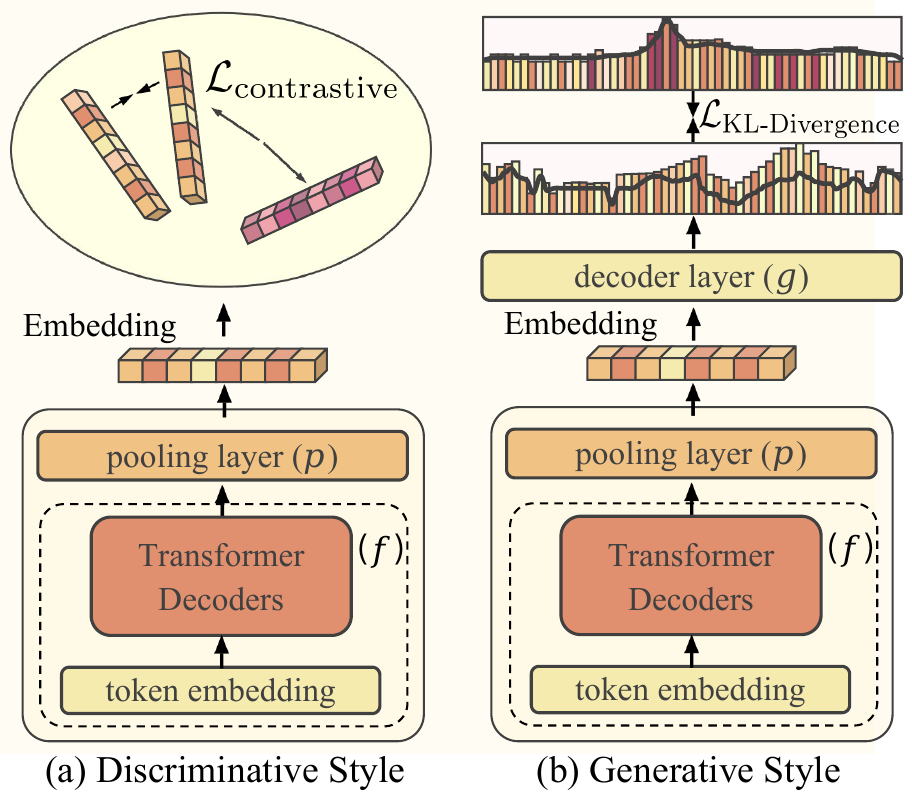} 
    \caption{Comparison between (a) traditional discriminative contrastive learning and (b) our proposed generative unsupervised framework: Text2Token.} 
    \label{fig:introduction}
\end{figure}

\section{Introduction}
Text representation learning (TRL) has demonstrated substantial progress, enabling the web to understand and process textual content semantically, improving search and recommendation, etc. However, text readily available on the web is often unlabeled, making it crucial to design efficient unsupervised methods that autonomously learning high-quality text representations from the web data.

With the rapid advancement of large language models (LLMs), embedders utilizing LLMs as backbones have demonstrated unprecedented generalization on TRL~\cite{muennighoff2023mteb}. However, most methods rely on contrastive learning over large-scale annotated or generated datasets, with their successful experience primarily stemming from the pretrained masked language model (PMLM) era \cite{reimers2019sentence,gao2021simcse}. As shown in Figure \ref{fig:introduction}(a), contrastive learning pulls positive examples closer and pushes negative examples further apart in the representation space, representing a typical discriminative-style approach. In contrast, as generative models, LLMs are trained using generative tasks throughout all stages. If a generative-style task can be used to train text representations, it would better leverage the capabilities learned by LLMs. Therefore, we question: Can a simple but effective generative approach be proposed for text representation learning? 

Before contrastive learning, generative methods, such as Skip-Thought~\cite{kiros2015skip} and FastSent~\cite{hill2016learning}, had dominated the unsupervised TRL field. However, subsequent attempts~\cite{wang2021tsdae,wu2022sentence} using generative methods failed to demonstrate superior performance compared to the simple contrastive learning method like SimCSE~\cite{gao2021simcse}. Recent TRL works on generative tasks \cite{gao2022unsupervised,xiao2022retromae,behnamghaderllm2vec,li2024llama2vec} primarily employ custom generative tasks as pretext tasks for contrastive learning, thereby enhancing the capabilities of contrastive fine-tuning. Even after the emergence of LLMs, no more natural generative task has been proposed to replace contrastive learning in TRL.

\begin{figure}[t]
    \centering
    \includegraphics[width=\linewidth]{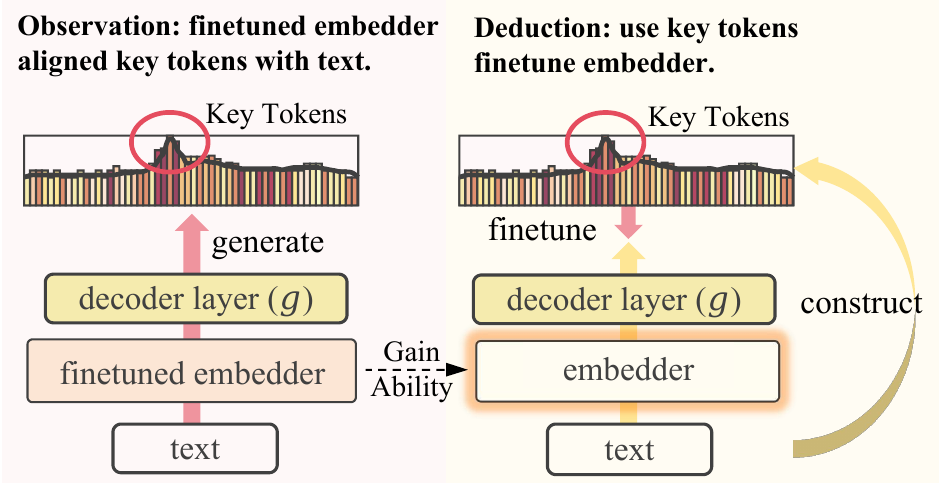} 
    \caption{The relation between the findings in \cite{nie2025text} (left) and the new proposed training method in this work (right).} 
    \label{fig:motivation}
\end{figure}

Obviously, the decline of generative tasks stems primarily from the scope of the supervision signals: the supervisory signal for the generative task originates from the discrete vocabulary space, while that for the discriminative (contrastive) task comes from the continuous representation space, directly affecting representation. For a long time, the community lacked an understanding of the relationship between continuous representation spaces and discrete lexical spaces. Fortunately, a recent empirical study \cite{nie2025text} finds the interesting relationship between the high-quality representations produced by LLM-based embedders and the vocabulary space. That is, when the representation output from the fine-tuned contrastive learning embedders is passed through the decoder layer of LLMs, the tokens with higher decoding probabilities, called the aligned tokens, are almost the key tokens of the input text. This phenomenon has been demonstrated to be prevalent across all LLM-based embedders. Interestingly, the decoder layer retains the original backbone parameters of the LLM and remains invisible during contrastive learning. The reason behind related to the simple variation of the principal components in the feature space during contrastive learning (See Section \ref{sec:analysis} for details). In this work, we primarily leverage this phenomenon to design new generative objectives, without delving into a deeper study of the underlying contrastive learning mechanisms. Specifically, empirical observations in \cite{nie2025text} demonstrate that representations trained with contrastive learning ``align'' with key tokens. As shown in Figure \ref{fig:motivation}, we hypothesize that this phenomenon is reversible: if training an embedder to generate key tokens, it to produce high-quality representations.

Inspired by this thought process, we propose a novel unsupervised learning framework, Text2Token, for TRL. As illustrated in Figure \ref{fig:introduction}(b), Text2Token utilizes the decoder layer of the LLM to map representations onto the vocabulary space, while receiving the KL-divergence supervisory guidance from a target token distribution (which we also refer to as the \textbf{token target}). When designing token targets, we do not rely on annotated data or external models, but solely utilize the statistical metrics of the dataset or intrinsic knowledge within the LLM backbone. Consequently, the training methodology is entirely unsupervised. Specifically, we first examined how high-quality representations ``align'' with the key tokens, identifying two essential categories of key tokens: (1) the core tokens from the nouns and adjectives in the text, and (2) semantically related tokens, including morphological variants and near-synonyms. We subsequently devised two distinct approaches to simulate these characteristics: (1) the data-driven method selects key tokens from the original text; (2) the model-derived method filters semantically relevant tokens from LLM priors. The final framework employs a two-stage training paradigm, successively utilizing data-driven and model-derived methods to obtain token targets as the target distribution in the KL-divergence loss.

Unlike the previous methods, Text2Token is intended to replace contrastive learning, instead of being a preliminary pretext task for contrastive learning. Therefore, we compared Text2Token with the state-of-the-art unsupervised TRL model, LLM2Vec \cite{behnamghaderllm2vec}. The performance on MTEB v2 \cite{muennighoff2023mteb} demonstrates that Text2Token achieves significant advantages across multiple tasks, substantially outperforming LLM2Vec in the average score. Additionally, as a new framework, we conducted extensive analytical experiments to aid in understanding the role of each component within the method and the impact of hyperparameters. Similarly, we compared our trained embedder with the contrastive fine-tuned embedder on both principal component analysis and token-level statistics, demonstrating that they achieve similar parameter solutions to a certain extent.

The main contributions of this paper are as follows:
\begin{itemize}
    \item  We propose a novel training framework for unsupervised TRL, Text2Token, that relies entirely on the generation objective, token target prediction, and differs fundamentally from all traditional methods based on contrastive learning.

    \item We explore two unsupervised approaches for constructing token distribution targets: a data-driven method and a model-derived method. Both approaches are integrated into our proposed training framework.

    \item Our embedder achieves superior performance on the MTEB benchmark compared with the unsupervised SOTA.

\end{itemize}

\section{Background}

\subsection{Unsupervised Text Representation Learning}
\paragraph{Embedding} Built upon the pretrained masked language models (PMLMs), such as BERT~\cite{devlin2019bert} and RoBERTa~\cite{liu2019roberta}, it effectively addresses the representation collapse that existed in earlier BERT-based models. Specifically, given a $n$-sized unlabeled corpus $\mathcal{D} = \{x_1, x_2, \ldots, x_n\}$, where $x_i$ denotes a text. Each text $x_i$ is represented as a token sequence $x_i = (w_1, w_2, \ldots, w_T)$, where $w_k$ denotes the $k$-th token and $T$ denotes the sequence length of $x_i$. When adapting the PMLM $F$ for text representation encoder $F'$, the masked language modeling head $g$ can be conceptually replaced by a pooling layer $p$ to produce text-level representations $e_i$, while preserving the rest module $f$ of $F$. The following is the interpretation in formula form:
\begin{align}
\label{formula1}
    F = g \circ f \to F' = p\circ f.
\end{align}

The text representation $\ve_i$ of a text $x_i$ is obtained as:
\begin{align}
\label{formula2}
\ve_i = p\bigl(\mathbf{h_i}) = p\bigl(f(x_i)\bigr),
\end{align}
where $\mathbf{h_i}$ denotes the last hidden states from $f$ and the layer $p(\cdot)$ can be mean pooling:
\begin{align}
\label{formula3}
\ve_i = \frac{1}{t_i} \sum_{k=1}^{t_i} \mathbf{h}_i^{k},
\end{align}
or using the representation of the \texttt{[CLS]} token:
\begin{align}
    \ve_i = \mathbf{h}_i^{[\texttt{CLS}]}.
\end{align}

\paragraph{Training} Contrastive learning \cite{oord2018representation,gao2021simcse,izacard2021unsupervised} represents a breakthrough in the field of unsupervised text representation learning in recent years. Its core lies in constructing the positive example pairs. For example, SimCSE \cite{gao2021simcse} obtains two different representations of each text $x_i$ with randomness of dropout masks; while Contriever \cite{izacard2021unsupervised} randomly crops two segments from the same document, performs random word deletion on them, and uses them as positive sample pairs. Then, each positive example pairs are encoded as the representation pair $(\ve_i, \ve_i^{(+)})$. The remaining representations within the same batch (with batch size $N$) are regarded as negative samples. The InfoNCE loss \cite{oord2018representation} for contrastive learning:
\begin{equation}\label{eqn:cl}
\small
    \mathcal{L}_{\rm cl} = - \frac{1}{N} \sum_{i=1}^{N} \log \frac{\exp\left(\mathrm{sim}\left(\ve_i, \ve_i^{(+)}\right) / \tau\right)}{\sum_{j=1}^{N} \exp\left(\mathrm{sim}\left(\ve_i, \ve_j^{(+)}\right) / \tau\right)},
\end{equation}
where $\mathrm{sim}(\cdot,\cdot)$ denotes the cosine similarity and $\tau$ is a temperature hyper-parameter. Thus, this training process requires no manual labels for the texts, and as a result, numerous methods have emerged to further improve SimCSE, including alternative positive sample construction~\cite{liu-etal-2021-fast, jiang-etal-2022-promptbert, wu-etal-2022-pcl,wu-etal-2022-esimcse}, hard negative mining~\cite{chen-etal-2023-alleviating, deng2023clustering, cao-etal-2022-exploring, wu-etal-2022-smoothed, zhou-etal-2022-debiased,Zhang_Zhang_Mensah_Liu_Mao_2022}, and loss function enhancements~\cite{zhang-etal-2022-contrastive, liu-etal-2023-rankcse, chen-etal-2022-information,chuang-etal-2022-diffcse}.

\paragraph{Improvements for LLMs} When the backbone model $F$ transitions to the LLM, embedding and optimization methods have not undergone significant changes. As mentioned in Eqn.~\ref{formula1} and Eqn.~\ref{formula2}, the embedder based on LLMs shares a similar formulation with the PMLMs, except that $g$ denotes the decoder layer, and $f$ denotes the remaining part. 
To better suit text representation tasks, bidirectional attention is enabled in $f$, following the design proposed by BeLLM~\cite{li-li-2024-bellm} and adopted in subsequent studies~\cite{zhao-etal-2025-prompt, muennighoff2024generative,behnamghaderllm2vec}.
In addition to retaining mean pooling as in Eqn.~\ref{formula3}, a novel prompt-based pooling strategy \cite{jiang2024scaling} tailored for LLMs is also employed. Specifically, a prompt $x_{\rm prompt}$ is prepended to guide semantic aggregation to the last token. For example, promptEOL \cite{jiang2024scaling} use the prompt \texttt{This sentence: ``[text]'' means in one word:``}, and the pooling strategy is expressed as
\begin{align}
   \ve_i = p(f(x_{\rm prompt}(x_i)) = \mathbf{h}_i^\text{last},
\label{formula4}
\end{align}
where $x_{\rm prompt}(\cdot)$ is the operation of replacing \texttt{[text]} with the real text, and $\mathbf{h}_i^{\rm last}$ represents the last hidden state of the last token.

Owing to the structural similarity between the LLMs and the PMLMs, contrastive learning continues to be the dominant training paradigm for text representation learning. Given a pair of texts with either a positive or negative relation, the model is trained to distinguish between similar and dissimilar pairs, which we regard as a discriminative training strategy.

\begin{figure*}[tbp]
    \centering
    \includegraphics[width=1\textwidth]{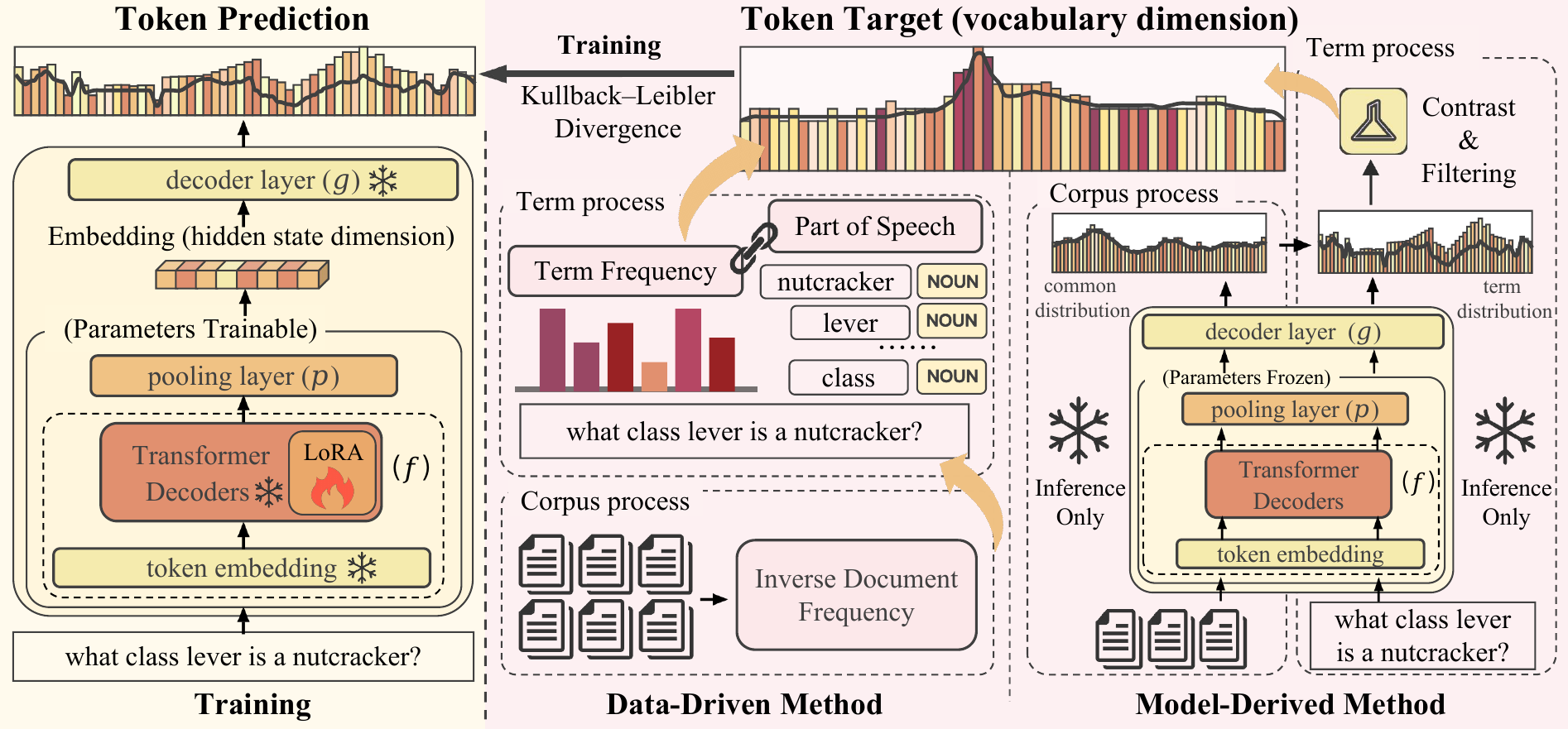} 
    \caption{The overview of our generative framework, Text2Token, for unsupervised text representation learning.} 
    \label{fig:main}
\end{figure*}

\subsection{Generative Methods for Text Representation}

Before contrastive learning emerged, some methods learn text representations through generative unsupervised tasks. For instance, Skip-Thought \cite{kiros2015skip} and FastSent \cite{hill2016learning} predict context sentences from the representation of the current text, while SDAE \cite{hill2016learning} added noise to the text and then reconstructed the original sentence using the representations. However, these methods performed less effectively than contrastive learning and are no longer mainstream paradigms.

In fact, following the emergence of contrastive learning, generative self-supervised tasks have predominantly appeared as the pretext tasks \cite{gao2022unsupervised,xiao2022retromae,behnamghaderllm2vec,li2024llama2vec}. Specifically, these works fine-tune the embedders with the pretext task (a designed generative unsupervised task) and the additional decoder module first, thereby enhancing the PLM's ability to adapt to aggregated semantics. Then, the fine-tuned embedders will undergo further fine-tuning through contrastive learning, showing better representation quality than those that are not fine-tuned by pretext tasks. The success of these methods implies that the optimization objectives for a certain class of generative tasks align with contrastive learning. However, these successful empirical cases alone do not enable us to derive effective general principles.

Fortunately, a recent study~\cite{nie2025text} shows that, if the original decoder layer $g$ of LLM is reattached after the embedder, even embedders trained with contrastive learning, the representations can effectively map the key tokens strongly aligned with the input text semantics. They define the generative structure $G$ as
\begin{equation}
    G = g \circ p \circ f, 
\label{eqn:training_model}
\end{equation}
Notably, the decoder layer $g$ is merely a simple linear transformation and is not involved in the finetuning of the embedder. In other words, the actual generative ability lies in the embedder itself. A well-trained embedder is thus capable of not only generating meaningful text representations but also generating key tokens. 
Then, conversely, could using key token prediction as the training task yield a high-quality embedder?
It suggests that text representation learning based on the generative paradigm may have a much broader development potential than we previously thought.

\section{Text2Token: Learning Framework}
\label{Learn From Token Target}

\subsection{Motivation}

Generative training paradigms have a natural compatibility with LLMs. However, the current mainstream approaches in text representation learning remain discriminative, such as contrastive learning, which limits the potential of LLMs in this field. From the study by \cite{nie2025text}, we observe that a well-trained embedder actually has a strong ability to generate key tokens of the input. Motivated by this, we hypothesize that \textbf{training an embedder to generate key tokens can also enable it to produce high-quality representations}, as illustrated on the right of Figure \ref{fig:motivation}. During training, the target distribution of vocabulary space guides the changes in the prediction distribution, influencing the representation space before the decoder layer $g$. Ultimately, both vocabulary and representation spaces converge to the optimal solution together.

\subsection{Overview}
Based on the above hypothesis, we propose a novel generative framework, Text2Token, for unsupervised text representation learning. Specifically, during training, Text2Token uses $G$ in Eqn.\ref{eqn:training_model} to predict the probability distribution of key tokens, supervised by a precomputed target key token probability distribution. After training, $g$ is discarded, and $F' = p \circ f$ serves as the embedder to obtain text representations. Clearly, the core of Text2Token is to define the target token distribution for each text, referred to as the \textbf{token target}. We reserve this core component for Section \ref{sec:token_target}, where we detail how we construct the token target through empirical research and insights from prior work. The remainder of this section will formally introduce our training strategy. Without additional explanation, the default notation for $G$'s backbone is derived from LLMs, consistent with the experimental section.

\subsection{Training Strategy}

\subsubsection{Token Prediction}
\label{sec:token_generation}
For each text $x_i$, we can obtain the vocabulary-size logit $\vs \in \mathbb{R}^{|\mathcal{V}|}$ through $G$:

\begin{align}\label{eq:lm_head}
\vs = G(x_i) = g(p(f(x_i)) = g(\ve_i) = \mathbf{W}_{\text{lm}}\ve_i,
\end{align}
where $\mathbf{W}_{\text{lm}} \in \mathbb{R}^{|\mathcal{V}| \times d}$ is the pre-trained parameter of $g$\footnote{To our best knowledge, the mainstream LLMs all follow the original design of GPT \cite{radford2018improving}, employing an unbiased matrix as the sole parameter for the decoder layer.} and remain fixed during training. Then, a softmax function is applied on $\vs$ to obtain the prediction distribution in the vocabulary space. For any $w_k$ in $\mathcal{V}$, its probability can be expressed as
\begin{align}
Q_{\theta}(w_k|x_i) &= \text{Softmax}(\vs)_k = \frac{\exp(\vs_k)}{\sum_{j=1}^{|\mathcal{V}|} \exp(\vs_j)}. \label{eq:vocab_scores}
\end{align}

\subsubsection{Token Target Construction}
\label{sec:token_target_construction}
For each  $x$, we generate a token target $P_{\text{target}}(w|x)$ over the model vocabulary $V$ to serve as the training objective. We refer to this token distribution as the \textbf{token target}. In our expectation, it should assign high probabilities to key tokens and low probabilities to non-key tokens. The token target construction process can be formulated as follows:

\begin{align}
\small
P_{\text{target}}(w_k|x_i) = \left\{
\begin{aligned}
& Con(x_i, \mathcal{D}),&\quad \text{\# data-driven}\\
& Con(x_i, \mathcal{D},\theta_0), &\quad \text{\# model-derived} \\
\end{aligned}
\right.
\end{align}
where the construction process of token target is denoted as $Con(\cdot)$ and $\theta_0$ is the pre-trained parameter of $G$. In Section \ref{sec:token_target}, we propose two $Con$ function methods: (1) the data-driven method relies solely on the training dataset $\mathcal{D}$; (2) the model-derived method additionally leverages the original model $\theta_0$, which plays a decisive role in shaping the target.

\begin{table*}[htbp]
\centering
    \small
   
    \begin{tabular}{@{} >{\centering\arraybackslash}m{0.10\linewidth} |>{\centering\arraybackslash}m{0.15\linewidth} |>{\centering\arraybackslash}m{0.65\linewidth} @{}}
    \toprule
    \textbf{BackBone} & \textbf{Model} & \textbf{Top 10 Aligned Tokens} \\
    \midrule
     &LLM2Vec-unsup &
      \tok{myred!70}{\_drum}\;\tok{myello!70}{\_No}\;\tok{myred!70}{\_tamb}\;\tok{myred!70}{\_drums}\;\tok{myred!70}{No}\;\tok{mygrey!70}{\_reson}\;\tok{mygrey!70}{\_tun}\;\tok{myello!70}{\_instrument}\;\tok{myred!70}{\_noise}\;\tok{myello!70}{\_frame} \\

    Mistral-7B &LLM2Vec-sup &
      \tok{myello!70}{\_instrument}\;\tok{myello!70}{\_No}\;\tok{myello!70}{\_per}\;\tok{myred!70}{\_instruments}\;\tok{myred!70}{\_Nor}\;\tok{myello!70}{\_string}\;\tok{myred!70}{\_musical}\;\tok{myello!70}{\_frame}\;\tok{myred!70}{\_}\;\tok{myred!70}{\_no} \\

    &GritLM &
      \tok{myred!70}{\_Nor}\;\tok{myello!70}{\_No}\;\tok{myred!70}{\_no}\;\tok{myello!70}{\_go}\;\tok{myred!70}{\_nor}\;\tok{myred!70}{\_instruments}\;\tok{myello!70}{\_instrument}\;\tok{myello!70}{\_string}\;\tok{myello!70}{\_per}\;\tok{myred!70}{\_gu} \\
      \midrule
   
    \multirow{2}{*}{Lllama3-8B} &LLM2Vec-unsup &
      \tok{myello!70}{\_instrument}\;\tok{myello!70}{\_perc}\;\tok{myello!70}{\_frame}\;\tok{myello!70}{\_guitar}\;\tok{myello!70}{\_like}\;\tok{myred!70}{\_finger}\;\tok{myred!70}{er}\;\tok{myello!70}{\_covered}\;\tok{mygrey!70}{icit}\;\tok{mygrey!70}{erc} \\

    &LLM2Vec-sup &
      \tok{myred!70}{\_Nor}\;\tok{myred!70}{Nor}\;\tok{myello!70}{\_instrument}\;\tok{myello!70}{\_string}\;\tok{myred!70}{14}\;\tok{myred!70}{\_nor}\;\tok{myello!70}{\_frame}\;\tok{myred!70}{\_instruments}\;\tok{myred!70}{\_Norris}\;\tok{myred!70}{nor}\ \\
    \midrule
    Llama2-7B &Llama2vec &
      \tok{myello!70}{\_No}\;\tok{myello!70}{\_like}\;\tok{myello!70}{\_string}\;\tok{myello!70}{\_instrument}\;\tok{myred!70}{\_what}\;\tok{myred!70}{\_}\;\tok{mygrey!70}{?}\;\tok{mygrey!70}{\_is}\;\tok{myello!70}{\_go}\;\tok{myred!70}{\_instruments} \\
    \bottomrule
  \end{tabular}

  \centering
    \caption{Top-10 Aligned Tokens from the different embedders for the text ``Noori is a 14-stringed instrument shaped like a guitar, but with a wooden frame covered in goatskin to produce percussive sounds like those of a djembe.''}
    \label{tab:section4-sota}
\end{table*}
\subsubsection{Loss Function}
To train the embedder $f$, we use $P_{\text{target}}(w|x)$ constructed in Section~\ref{sec:token_target_construction} as the training objective and optimize $P_{\text{aligned}}(w|x)$ from Section~\ref{sec:token_generation} by minimizing the KL divergence:
\begin{equation}\label{eqn:kl_loss}
\mathcal{L}_{\rm text2token} = \text{KL}\left(P_{\text{target}} \| Q_{\theta}\right).
\end{equation}

\section{Token Target Design}\label{sec:token_target}

\subsection{Observation and Discussion}
\label{sec:observation}

Recall that Text2Token is motivated by the empirical observation that the representations produced by the LLM-based embedders align with key tokens, although they are fine-tuned by contrastive learning \cite{nie2025text}. Therefore, we start by analyzing several state-of-the-art embedders to examine the aligned token distributions mapped from their text representations. This analysis allows us to investigate further how a high-quality representation aligns with an appropriate token distribution. The reference models studied are:
\begin{itemize}
    \item \textbf{LLM2Vec-unsup~\cite{behnamghaderllm2vec}}.
    LLM2Vec converts decoder-only LLMs into text encoders via three unsupervised steps: enable bidirectional attention, brief masked next-token adaptation, and SimCSE-style contrastive learning.
    \item \textbf{LLM2Vec-sup~\cite{behnamghaderllm2vec}}. The supervised version of LLM2Vec.
    \item \textbf{GritLM~\cite{muennighoff2024generative}}. GritLM is a single instruction-tuned LLM that jointly trains generative and contrastive objectives, toggled by task instructions.
    \item \textbf{Llama2vec~\cite{li2024llama2vec}}. Llama2Vec adapts LLMs for dense retrieval via two unsupervised pretext tasks to learn text representations: EBAE (self-reconstruction) and EBAR (next-sentence prediction). 
\end{itemize}

As illustrated in Table~\ref{tab:section4-sota}, although different models align to different tokens, these top-aligned tokens share the following common characteristics: (1) {\bf the key tokens appearing in the text (highlighted in \tok{myello!70}{yellow}) generally receive high probabilities}. For example, ``\_No'' (the first token of the proper noun ``Noori''), ``\_instrument'' and ``\_string'' Tokens sharing the same stem with in-text tokens also exhibit elevated probabilities, such as ``\_instruments'' and ``\_Nor''; (2) \textbf{the semantically-related tokens that do not occur in the surface text (highlighted in \tok{myred!70}{red}) are often assigned high probabilities by some embedders}. For instance, LLM2Vec-unsup aligns the text with ``drum'' consistent with the phrase ``sounds like those of a djembe'' where ``djembe'' refers to a type of drum. Similarly, LLM2Vec-sup maps it to ``\_musical'', and Llama2vec aligns it to ``what'', which is reasonable given that the text introduces the instrument ``Noori''; (3) \textbf{Only a very small number of unrelated tokens (highlighted in \tok{mygrey!70}{grey}) are assigned high probabilities by some embedders.} More examples are shown in Appendix~\ref{sec:further_study}.

The above discussion provides a foundation for independently constructing token targets. Our goal is to design token targets that closely capture the beneficial characteristics of these aligned distributions. As shown in the Figure~\ref{fig:main} on the right, we designed two constructors to simulate the in-text token distribution (the data-driven Method) and out-of-text token distribution (the model-derived Method) in aligned token distributions, respectively. Then use them in sequence to guide the training. The specific details of each method differ and are described in the next two sections.

\subsection{Data-Driven Method}

From the preceding analysis, we observe that the key token in the original text constitutes a crucial component of the aligned token target. In addition, Llama2vec~\cite{li2024llama2vec} introduced EBAE, a pretext task that attempts to recover all tokens appearing in the original text. However, EBAE can only serve as a pretext task before contrastive learning fine-tuning; the embedder trained solely using it exhibits suboptimal performance (Shown in Table \ref{tab:main_result}). We therefore propose constructing a token target based on these key tokens to approximate the capabilities of the aligned token target and achieve comparable training effects.

Intuitively, we should assign a significance score to each token in the text, reflecting its contextual salience. The key tokens are then selected according to these scores, and their scores are reflected as weights in the resulting distribution. To identify core words, we employ statistical NLP techniques due to their simplicity and effectiveness. Specifically, we adopt TF-IDF \cite{ramos2003using} for scoring and part-of-speech (POS) for further filtering. The POS filter is motivated by the previous empirical study \cite{nie2025text}, which finds that the majority of tokens aligned by the trained embedders are Noun (NOUN), Proper Noun (PNOUN), and Adjective (ADJ). The complete target-construction pipeline is illustrated below.

\subsubsection{Corpus Process}
To capture global information, the inverse document frequency (IDF) of each word $w_k$ is computed as
\begin{align}
    \text{IDF}(w_k) = \log\left(\frac{|\mathcal{D}|}{|\{x \in \mathcal{D} | w_k \in x\}|}\right),
\end{align}
where $|\cdot|$ is the cardinality operation on a set. The set $\{x\in \mathcal{D} | w_k \in x\}$ contains all documents in which appears at least once, and its cardinality gives the document frequency of $w_i$.

\subsubsection{Term Process}

For a document $x_i \in \mathcal{D}$, the TF-IDF score of token $w_k$ in $x_i$ is calculated as follows:
\begin{align}
    \text{TF}(w_k| x_i) = \frac{\text{count}(w_k | x_i)}{\sum_{j} \text{count}(w_k| x_j)},
\end{align}

\begin{align}
\text{Score}_{\text{tf-idf}}(w_k | x_i) = \text{TF}(w_k | x_i) * \text{IDF}(w_k),
\end{align}
where $\text{TF}(w_k | x_i)$ represents the term frequency. Furthermore, we use POS to conduct a further screening of the core words. Inspired by empirical observation of \cite{nie2025text}, we set up the POS filter set as
\[
\mathcal{P} = \{\text{NOUN},\text{PROPN},\text{ADJ}\}.
\]

The filtering process is as follows:
\begin{align}
\small
\text{Score}_{\text{pos}}(w_k | x_i)
= \mathbb{I} \left[\operatorname{pos}(w_k | x_i)\in\mathcal{P}\right] 
* \text{Score}_{\text{tf-idf}}(w_k | x_i)
\end{align}
where $\mathbb{I}[\cdot]$ is the indicator function and ${\rm pos}(\cdot)$ is the function to return the part of speech of $w_k$ in the context of $x_i$. \footnote{We utilize \texttt{en\_core\_web\_sm} model provided by spaCy \cite{spacy2020} for word-level part-of-speech tagging and assign the part-of-speech tag of each word to its split tokens.} Finally, we constructed token target based on the $\text{Score}_{\text{pos}}$:
\begin{align}\label{eqn:data_driven}
\small
    P_{\text{target}}(w_k | x_i) &= Con(x_i, \mathcal{D}) \\
    &= 
    \frac{\exp(\text{Score}_{\text{pos}}(w_k | x_i))}{\sum_{w_j \in \mathcal{V}} \exp(\text{Score}_{\text{pos}}(w_j | x_i))}.
\end{align}

\begin{table*}[ht]
\centering
\resizebox{\textwidth}{!}{
\begin{tabular}{ll|ccccccc|c}
\toprule
\multirow{2}{*}{\textbf{Backbone}} &\textbf{Categories →} & \textbf{Class.} & \textbf{Clust.}& \textbf{PairClass.}& \textbf{Rerank.}& \textbf{Retr.}& \textbf{STS}& \textbf{Summ.} &\textbf{Avg} \\
&\textbf{$\#$ of datasets →} &\textbf{8} & \textbf{8}& \textbf{3}& \textbf{2}& \textbf{10}& \textbf{9}& \textbf{1}& \textbf{41}\\
\midrule

\multirow{6}{*}{\textit{Llama3-8B}} &Llama2vec (EBAE)&66.93  & 37.76  &	65.46  &	38.56  &	7.86  &	56.27  &	8.31  &	40.16  \\
& LLM2Vec-unsup  & \textbf{76.25}  & 44.99  & \textbf{78.01} & 43.24 & 22.35  & \textbf{75.55} & \textbf{32.62} & 53.29 \\
&Text2Token (Last w causal) &70.98  & \textbf{51.51}  & \underline{76.09} & \textbf{45.60} & \textbf{41.57}  & \underline{72.86} & \underline{28.16} & \textbf{55.25} \\
&Text2Token (Last w bidirectional) & \underline{75.68}  & \underline{48.17}  &75.77 &	\underline{45.38} &	\underline{38.35} &	69.36 &	25.23 &	\underline{53.99} \\
&Text2Token (Mean w causal) &67.32 & 46.20 &	64.43 &	40.76 &	26.15 &	59.82 &	15.44 &	45.73 \\
&Text2Token (Mean w bidirectional) &66.94 &	44.14 &	59.69 &	36.77 &	8.67 &	52.99 &	9.40 &	39.80 \\

\midrule
\multirow{6}{*}{\textit{Mistral-7B}} 
&Llama2vec (EBAE)&68.37   & 36.02   &	66.24   &	38.28   &	9.40   &62.35   &	14.43   &	42.16  \\
& LLM2Vec-unsup  &\textbf{75.72} & 40.74 & \textbf{80.94} & 44.17 & 31.05 & \textbf{78.24} & \textbf{26.69} & \underline{53.94} \\
&Text2Token (Last w causal) & 71.65 & \textbf{49.71} & 76.98 &	\underline{44.85} &	\textbf{40.12} &	72.05 &	17.20 &	53.22 \\
&Text2Token (Last w bidirectional) & \underline{74.22} & \underline{49.08} &	\underline{78.61} &	\textbf{45.39} &	38.12 &	\underline{73.93} &	\underline{26.30} &	\textbf{55.10} \\
&Text2Token (Mean w causal)  &66.08 & 43.48 &	62.65 &	39.57 &	26.23 &	56.34 &	4.79 &	42.73 \\
&Text2Token (Mean w bidirectional) &71.51 & 48.56 &	75.61 &	44.57 &	\underline{39.86} &	71.89 &	10.05 &	51.72 \\
\bottomrule
\end{tabular}
}
\caption{Performance comparison with unsupervised state-of-art, LLM2Vec, on MTEB.v2 benchmark. The \textbf{bold} font indicates the best score in each category, while \underline{underlined} text indicates the second-best score. Detail results illustrated in Appendix~\ref{Detailed Results}.}
\label{tab:main_result}
\end{table*}
\subsection{Model-Derived Method}

Beyond the words explicitly present in the text, the aligned token target includes absent tokens that are morphological variants or semantically related terms. 
Such tokens cannot be extracted directly from the original text, but they can emerge through the decoder’s inherent inference capability. In our fine-tuned model, such key tokens already receive high weights. Whereas in the untuned model, high-weight tokens contain not only these key tokens but also various noisy elements.

Inspired by prior works on representative words prediction (ROP) \cite{ma2021prop,ma2021b} and contrastive decoding \cite{wingate2022prompt, lin2024critical}, we introduce a common token distribution to contrast and filter each text’s original token distribution. Specifically, the common distribution is defined as the mean of the token distributions mapped from multiple texts across the corpus. Tokens that consistently achieve high probabilities across all texts are less likely to be the key tokens of any particular text. Therefore, tokens with high probability in the common distribution can be removed from a text’s original token distribution. The final filtered distribution serves as the training target to guide the model’s own learning. The full pipeline for constructing this target is illustrated below.

\subsubsection{Corpus process}
After weighing the inference cost and training efficiency, we first estimate the common token distribution by computing the expectation over a corpus subset $\mathcal{D}_{\rm subset}$ with a rational size:
\[
\overline{Q}_{\theta_0}(w_k) = \frac{1}{|\mathcal{D}_{\rm subset}|} \sum_{x \in \mathcal{D}_{\rm subset}} Q_{\theta_0} (w_k | x_i),
\]
where $\mathcal{D}_{\rm subset}$ denotes a sampled subset of the corpus.

\subsubsection{Term process}
We then compute the token distribution for each text using the same procedure. Then, we filter the token distribution $Q_{\theta_0}(w_k | x_i)$, using the common token distribution $\overline{Q}_{\theta_0}(w_k)$ as a reference, employing a relative confidence screening method to obtain the final token targets. We compare the performance of different filtering formulas in Appendix \ref{sec:filter_comparsion}, and the best one is
\begin{equation}
\hat{y}_{w_k} = 
\frac{Q_{\theta_0}(w_k | x_i)}{Q_{\theta_0}(w_k | x_i) + \overline{Q}_{\theta_0}(w_k)},
\end{equation}
\begin{equation}\label{eqn:model_derived}
\small
P_{\text{target}}(w_k | x_i) = Con(x_i, \mathcal{D},\theta_0) = 
\frac{\exp(\hat{y}_{w_k}/\tau)}{\sum_{w_j \in \mathcal{V}} \exp(\hat{y}_{w_j}/\tau)},
\end{equation}
where $\tau$ is the temperature hyper-parameter for adjusting the distribution smoothness, similar to InfoNCE Loss (Eqn.\ref{eqn:cl}).

\subsection{Two-Stage Training Paradigm}
Since the two targets emphasize different aspects of key tokens, we adopt a two-stage training strategy. In both stages, training is conducted using only the KL divergence (Eqn.\ref{eqn:kl_loss}). The token target in the first and second stage is from the data-driven method (Eqn.\ref{eqn:data_driven}) and the model-derived method (Eqn.\ref{eqn:model_derived}) by default. We further discuss the impact of varying the order in Section \ref{sec:ablation}.

\section{Experiment}
\subsection{Settings}
\subsubsection{Training}
Training was fully unsupervised, using the Wikipedia corpus preprocessed by DPR~\cite{karpukhin2020dense}. It proceeded in two stages, each with a different target as the training signal: 500 steps in the first stage and 200 steps in the second. We use LoRA with $r=16$, $\alpha=32$, on \texttt{q\_proj} and \texttt{v\_proj}, with 0.05 dropout and no bias update. The batch size is 64, and the sequence length is 128. Learning rate is 3e-5. The temperature $\tau$ and the constructing data size of the common distribution for the model-derived method are 0.0001 and 100K.

\subsubsection{Baseline}
We adopt LLM2Vec-unsup and Llama2Vec (EBAE) as our baselines. Both models employ unsupervised training approaches: LLM2Vec-unsup follows a contrastive learning paradigm, while Llama2Vec (EBAE) is based on a self-reconstruction objective. Unlike Text2Token, EBAE assigns equal probability to all tokens that appeared in the original text within the token target.

\subsubsection{Evalutaion}
Evaluation was performed on the Massive Text Embedding Benchmark (MTEB) \cite{muennighoff2023mteb}, using the English v2 release, covering 41 datasets in total. Ablations and analyses are conducted in a subset of MTEB presented in Table~\ref{tab:subset}.

\subsubsection{Technical Factors}
The experiments were conducted on two backbone models, Llama3-8B and Mistral-7B. We used ``causal'' to represent conventional attention and ``bidirectional'' to represent bidirectional attention. We explore two pooling strategies: (1) Mean: mean pooling (Eqn.\ref{formula3}), and (2) Last: last pooling (Eqn.\ref{formula4}).

\subsection{Main Results}
Table~\ref{tab:main_result} presents the results of our main experiment. To allow both categories of targets to contribute to the supervision signal, we adopt a two-stage training strategy in which each target type supervises one training stage. It can be observed that Text2Token with the last-pooling method outperforms LLM2Vec on average. This validates the effectiveness of our training framework. Specifically, Text2Token demonstrates consistent advantages across Clustering, Reranking, and Retrieval tasks. Moreover, we find that Text2Token achieves better results with last-pooling, whereas LLM2Vec is more suited to mean-pooling. \textbf{In summary, these results confirm that Text2Token achieves consistently superior performance across tasks, particularly with the last-pooling strategy.}

\subsection{Ablation Study}\label{sec:ablation}

Text2Token involves several adjustable factors, and ablation experiments are necessary to clarify the contribution of each component. Specifically, we first examine how different target types, pooling strategies, and attention mechanisms influence training performance for a single-stage setting. Further, we investigate whether the applied order of token targets affects the results significantly.

\begin{table}[tbp]
\centering
\resizebox{0.5\textwidth}{!}{
\begin{tabular}{lcc}
\toprule
\textbf{Stage I} &  \textbf{Stage II} & \textbf{Avg} \\
\midrule
\multicolumn{3}{c}{\textit{Llama3-8B} (w causal)} \\
\midrule
data-driven (Last) & model-derived (Last) & 61.57  \\
model-derived (Last) & data-driven (Last) & 61.58  \\
\midrule
data-driven (Mean) & model-derived (Mean) & 54.35  \\
model-derived (Mean) & data-driven (Mean) & 47.79  \\
\midrule
\multicolumn{3}{c}{\textit{Mistral-7B} (w bidirectional)} \\
\midrule
data-driven (Last) & model-derived (Last) & 59.96  \\
model-derived (Last) & data-driven (Last) & 55.99  \\
\midrule
data-driven (Mean) & model-derived (Mean) & 57.06  \\
model-derived (Mean) & data-driven (Mean) & 51.59  \\
\bottomrule
\end{tabular}
}
\caption{Ablation results of two-stage training.}
\label{tab:ablution2}
\end{table}
\subsubsection{Single-Stage Training}
\label{5.3.1}
\begin{figure}[t]
   \begin{subfigure}{\linewidth}
        \centering
        \includegraphics[width=\linewidth]{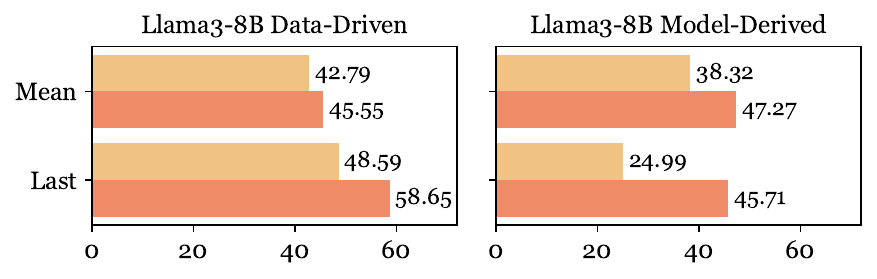}
    \end{subfigure}

    \begin{subfigure}{\linewidth}
        \centering
        \includegraphics[width=\linewidth]{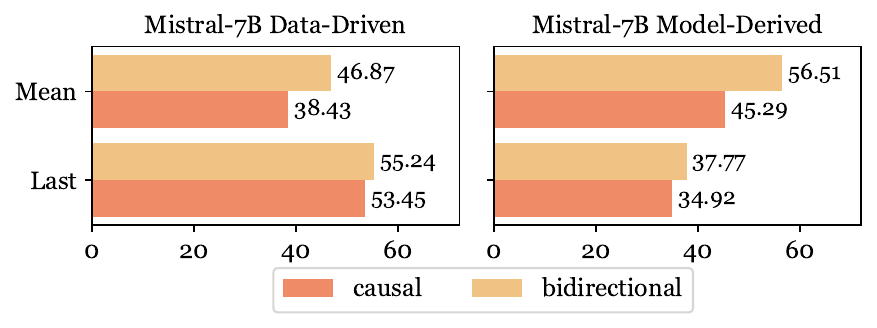}
    \end{subfigure}
    \caption{Ablation results of single-stage training.}
    \label{fig:ablation1}
\end{figure}

As shown in Figure~\ref{fig:ablation1}, we observe that data-driven targets achieve the best performance when paired with last pooling, while model-derived targets are more effective with mean pooling. Moreover, Llama3-8B benefits more from causal attention, whereas Mistral-7B performs better with bidirectional attention. These findings reveal clear complementarities between target types, pooling strategies, and attention mechanisms.

\begin{figure}[t]
    \centering
    \includegraphics[width=\linewidth]{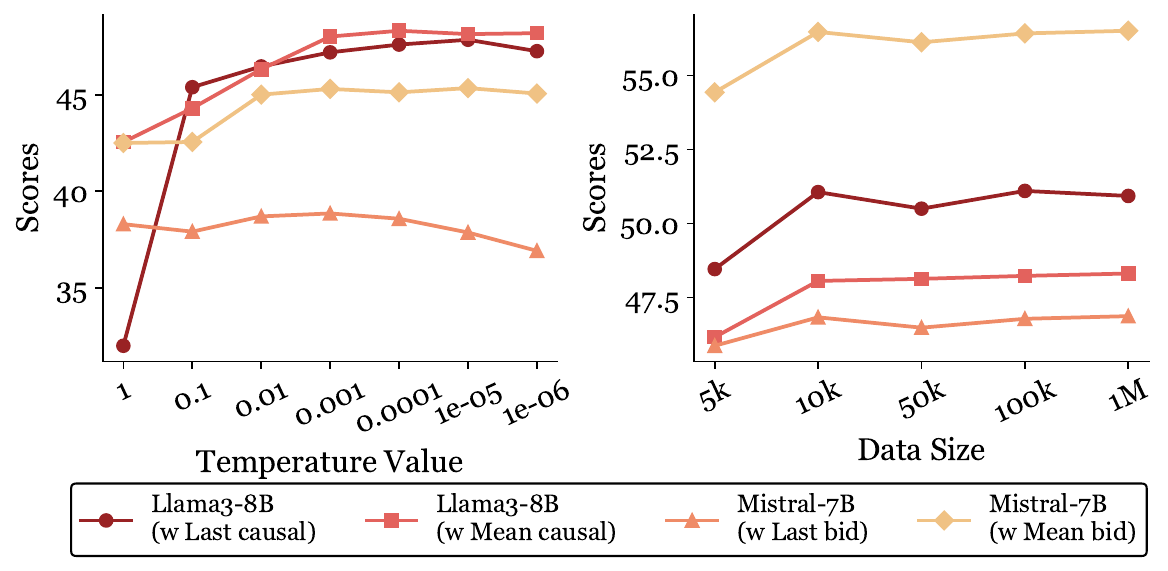}
    \caption{The result variation with the hyperparameter.}
    \label{fig:hyperparameter}
\end{figure}

\begin{table*}[htbp]
\centering
\small
    
    \begin{tabular}{@{} >{\centering\arraybackslash}m{0.2\linewidth} |     >{\centering\arraybackslash}m{0.75\linewidth} @{}} 
    \toprule
    \textbf{Model} & \textbf{Top 10 Aligned Tokens} \\
    \midrule
    Raw (Mistral-7B)&
      \tok{mygrey!70}{\textbackslash n}\;\tok{mygrey!70}{,}\;\tok{mygrey!70}{...}\;\tok{mygrey!70}{\_}\;\tok{myello!70}{\_a}\;\tok{mygrey!70}{\_and}\;\tok{myred!70}{\_strings}\;\tok{mygrey!70}{.}\;\tok{mygrey!70}{\_in}\;\tok{myred!70}{\_drum} \\
  
    Text2Token (Mistral-7B) &
      \tok{myred!70}{\_drum}\;\tok{myello!70}{\_frame}\;\tok{myred!70}{\_drums}\;\tok{myello!70}{\_instrument}\;\tok{myello!70}{\_No}\;\tok{myello!70}{\_guitar}\;\tok{myred!70}{\_strings}\;\tok{myred!70}{\_tamb}\;\tok{myello!70}{\_string}\;\tok{myred!70}{\_instruments} \\

    \midrule
    Raw (Llama3-8B) &
      \tok{myred!70}{\_No}\;\tok{mygrey!70}{\_K}\;\tok{myred!70}{\_Dj}\;\tok{myred!70}{\_Instrument}\;\tok{myello!70}{No}\;\tok{mygrey!70}{?}\;\tok{myred!70}{\_Viol}\;\tok{myred!70}{Instrument}\;\tok{mygrey!70}{K}\;\tok{myello!70}{\_instrument} \\
    Text2Token (Llama3-8B) &
      \tok{myello!70}{\_instrument}\;\tok{myred!70}{instrument}\;\tok{myred!70}{\_strings}\;\tok{myred!70}{strings}\;\tok{myred!70}{\_instruments}\;\tok{myello!70}{\_string}\;\tok{myello!70}{\_guitar}\;\tok{myello!70}{No}\;\tok{myred!70}{\_No}\;\tok{myello!70}{\_dj} \\
    \bottomrule
  \end{tabular}
  \centering
    \caption{Top-10 aligned tokens on the backbone and our trained embedders for the Sentence ``Noori is a 14-stringed instrument shaped like a guitar, but with a wooden frame covered in goatskin to produce percussive sounds like those of a djembe.''.}
    \label{fig:low}
\end{table*}

\begin{figure*}[htbp]
    \centering

    \begin{subfigure}{0.24\textwidth}
        \centering
        \includegraphics[width=\linewidth]{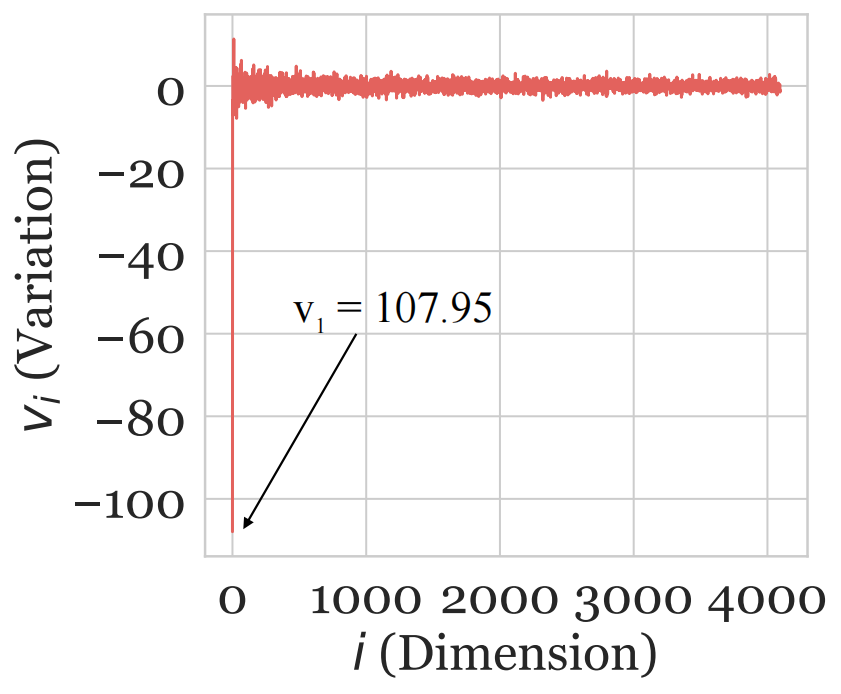}
        \caption{Llama3-8B $\rightarrow$ Text2Token}
    \end{subfigure}
    \hfill
    \begin{subfigure}{0.24\textwidth}
        \centering
        \includegraphics[width=\linewidth]{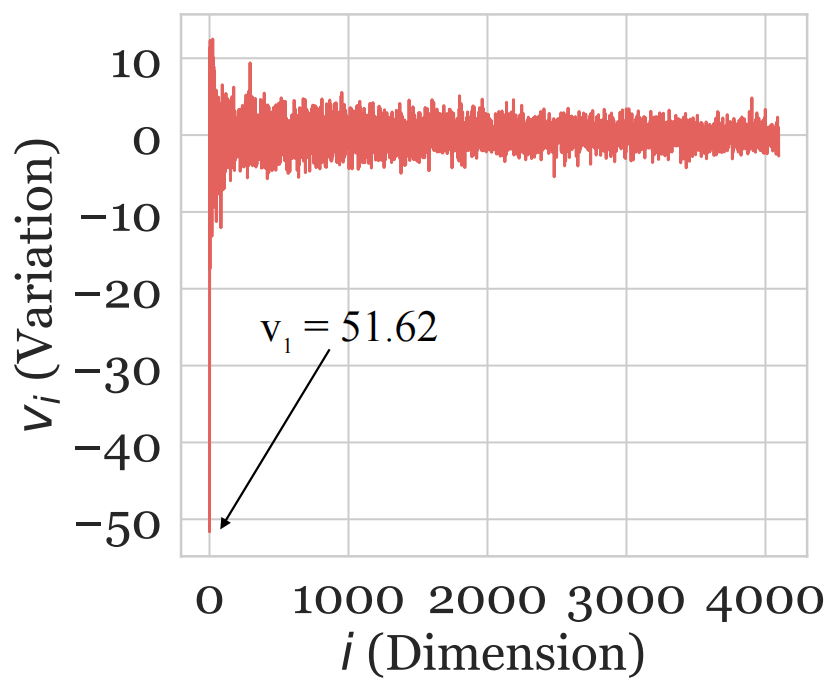}
        \caption{Llama3-8B $\rightarrow$ LLM2Vec}
    \end{subfigure}
  \hfill
    \begin{subfigure}{0.24\textwidth}
        \centering
        \includegraphics[width=\linewidth]{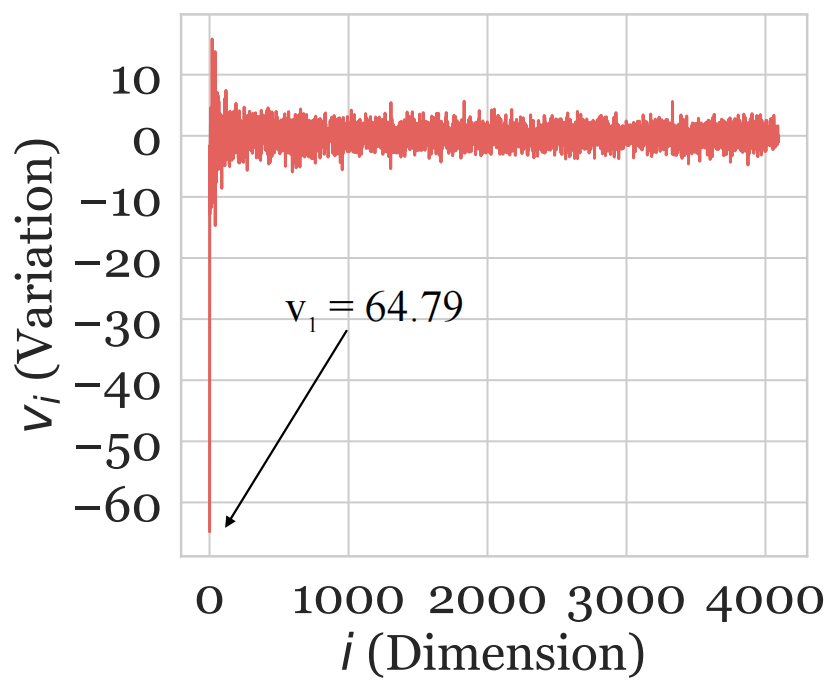}
        \caption{Mistral-7B $\rightarrow$ Text2Token}
        \label{fig:subfig3}
    \end{subfigure}
    \hfill
    \begin{subfigure}{0.24\textwidth}
        \centering
        \includegraphics[width=\linewidth]{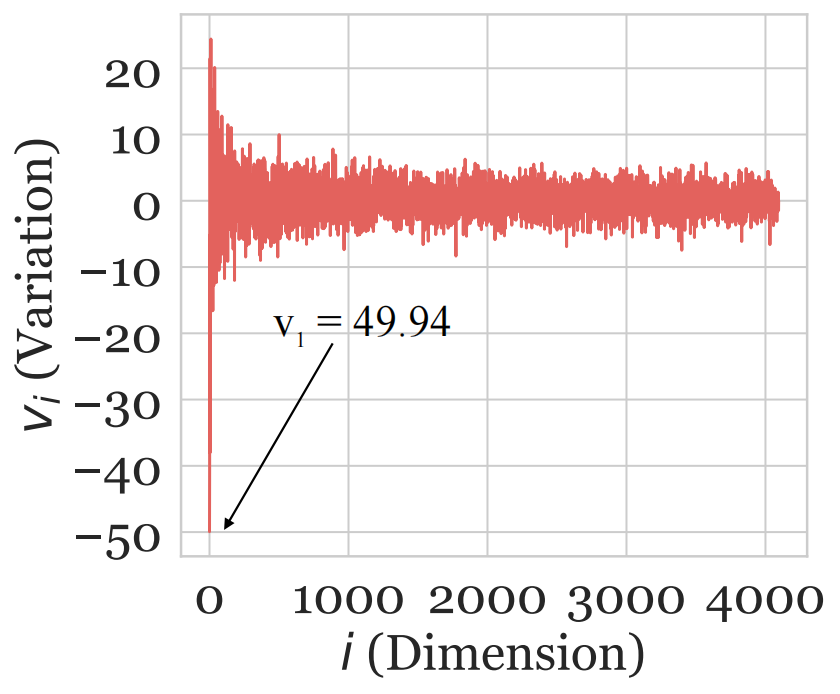}
        \caption{Mistral-7B $\rightarrow$ LLM2Vec}
        \label{fig:subfig4}
    \end{subfigure}

    \caption{Variation of Text2Token and Variation of LLM2Vec along each principal component of the representation space. $v_1$ means the variation of the first principal component.}
    \label{fig:high}
\end{figure*}

\subsubsection{Two-stage Training with Same Pooling}
\label{5.3.2}
As shown in Table~\ref{tab:ablution2}, we assign the two targets to the two stages and further verify whether their training order. Based on prior observations in the Section~\ref{5.3.1}, we use Llama3-8B with causal attention and Mistral-7B with bidirectional attention as backbones. First and foremost, it is crucial to emphasize that the experimental results can prove that two-stage training is indeed superior to that of the two single-stage trainings conducted separately. This validates the effectiveness of our approach of guiding the model with two types of targets in a two-stage training process. 
Secondly, the results also reveal that training with data-driven targets first generally yields slightly better performance, though in certain cases, such as the method of Llama3-8B with last pooling, the target order has minimal influence on training. 
These results confirm the clear advantages of the two-stage training strategy over the single-stage ones.

\subsection{Model-Derived Hyperparameter Study}
In this subsection, we discuss the influence of two hyperparameters, (1) temperature for distribution and (2) data size for common token distribution, when constructing the model-derived target.

\subsubsection{Temperature for Distribution}
In the model-derived method, a critical parameter is the temperature parameter $\tau$, which is used to smooth the final target distribution. We conducted experiments on Mistral-7B, and as illustrated in the figure~\ref{fig:hyperparameter} (left), under the same experimental settings, adjusting the temperature within the range from 0.01 to 1e-5 yields stable and consistent results, indicating that \textbf{the model-derived target is robust to its variations.}

\subsubsection{Data Size for Common Token Distribution}
Recall that the construction of the model-derived target depends on the LLM backbone's prior knowledge, and a common token distribution is needed to filter and obtain the token target. We evaluate how dataset size affects the quality of this distribution to reduce inference cost. Figure~\ref{fig:hyperparameter} (right) shows results using 5k, 10k, 50k, 100k, and 1M samples. Performance stabilizes beyond 10k samples, suggesting that \textbf{a relatively small dataset suffices to capture the macro-level distribution, while excessive data may risk overfitting.}

\subsection{Analysis}
\label{sec:analysis}
We examine the changes in Text2Token before and after fine-tuning, focusing on two aspects: (i) the variation of its correspondence with tokens in the vocabulary space, and (ii) the adjustments in the representation space. Although our approach differs from conventional contrastive learning, we further compare these changes with those of LLM2Vec to validate their consistency, showing that \textbf{Text2Token demonstrates well-aligned behavior during fine-tuning, exhibiting contrastive-like behavior in both token alignment and representation space variation.}

\subsubsection{Key Token Prediction}
As shown in the Table~\ref{fig:low}, after fine-tuning, the top tokens corresponding to the same text change significantly. These include not only in-text key tokens such as ``\_instrument'' and ``\_string'', but also out-of-text key tokens such as ``\_tamb'' and ``\_drums''. Moreover, when compared with the tokens predicted by other embedders reported in Table~\ref{tab:section4-sota}, Text2Token achieves a comparable or even higher level in predicting key tokens. More examples are shown in Appendix~\ref{Aligned Token}.

\subsubsection{Variation on Embedding Space}
The prior study~\cite{nie2025text} shows that the spectral variation of existing embedders (e.g., LLM2Vec) is primarily concentrated in the first principal component ($v_1$), which often captures noise in high-dimensional spaces. Thus, analyzing changes in $v_1$ provides a means to assess fine-tuning effectiveness. As shown in Figure~\ref{fig:high}, LLM2Vec follows this pattern, and Text2Token exhibits similar behavior, indicating that our training strategy effectively regularizes the high-dimensional space.

\section{Conclusion}
In this work, we introduced Text2Token to learn text representation with the generative objectives without supervision, moving beyond the conventional reliance on contrastive learning. By defining the aligned token target and analyzing its composition, we proposed two practical strategies for generating synthetic token targets: data-driven and model-derived. Text2Token achieves state-of-the-art results among unsupervised methods on MTEB v2. Importantly, we verified the connection between token distributions and representations during training, suggesting a new perspective for designing more effective generative TRL strategies in future.  

\bibliography{custom}

\clearpage
\appendix
\section{The Subset of MTEB.v2 Used for Analysis}
\label{subset}
Since the MTEB benchmark includes numerous tasks, my analysis experiments are also extensive. Therefore, we use a subset shown as Table~\ref{tab:subset} to improve efficiency. The subset ensures coverage of all task categories, maintaining the validity of the results.
\begin{table}[htbp]
\centering
\resizebox{0.35\textwidth}{!}{
\begin{tabular}{lc}
\toprule
\textbf{Category} & \textbf{Dataset} \\

\midrule
\multirow{2}{*}{Retrieval} & ArguAna  \\
 & FEVERHardNegatives.   \\
\midrule
Reranking & AskUbuntuDupQuestions  \\
\midrule
\multirow{3}{*}{Clustering} & BiorxivClusteringP2P.v2 \\
& MedrxivClusteringS2S.v2 \\
& TwentyNewsgroupsClustering.v2 \\
\midrule
PairClassification & SprintDuplicateQuestions  \\
\midrule
\multirow{2}{*}{Classification} & Banking77Classification \\
& MassiveIntentClassification \\
\midrule
\multirow{2}{*}{STS} & SICK-R \\
& STSBenchmark \\
\midrule
Summarization & SummEvalSummarization.v2 \\

\bottomrule
\end{tabular}
}
\caption{Subset of MTEB tasks used for ablations and analysis}
\label{tab:subset}
\end{table}

\section{Further Token Target Study}
\label{sec:further_study}

In Table~\ref{tab:section4-sota_plus}, we provide three more examples from
LLM2Vec, GritLM, and Llama2vec. We observe similar phenomena as in Section \ref{sec:observation}.

To make these observations more general, we performed a quantitative analysis. We randomly sampled 1,000 texts from Wikipedia, mapped each onto the vocabulary, and recorded whether their top-10 tokens occurred in the corresponding text. 

\begin{figure}[htbp]
\centering
    \centering
    \includegraphics[width=0.9\linewidth]{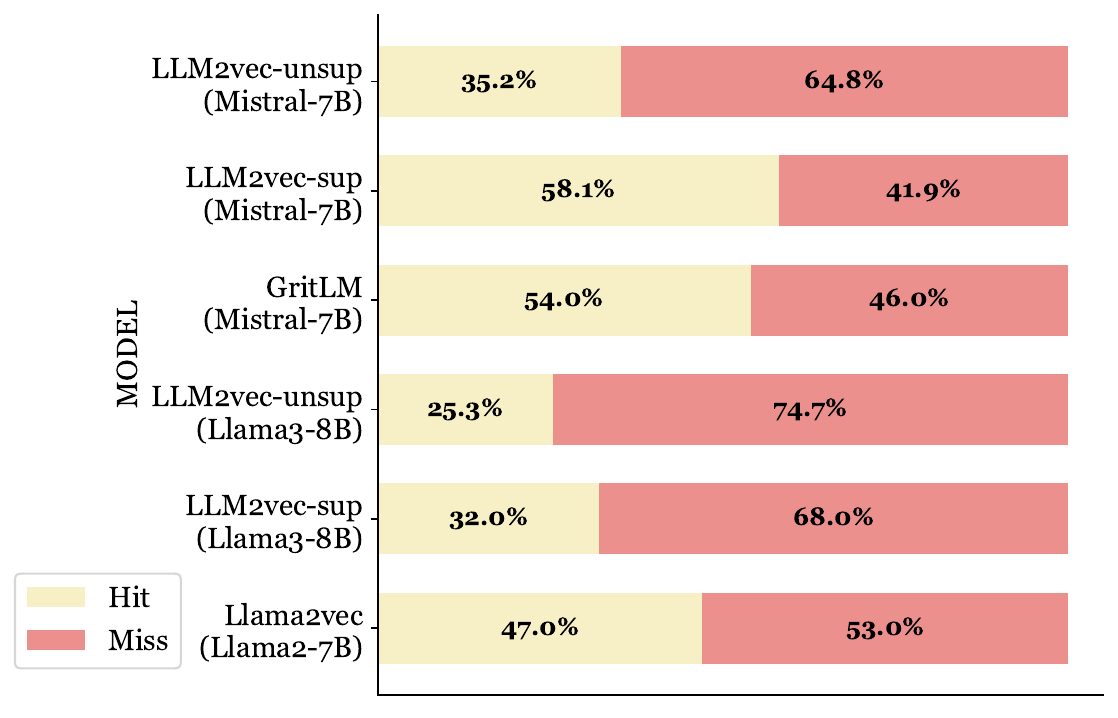}
    \caption{Cumulative Occurrence of Top-10 Tokens}
    \label{fig:section4-all}
\end{figure}

\begin{figure}[htbp]
    \centering
    \includegraphics[width=\linewidth,height=5cm,keepaspectratio]{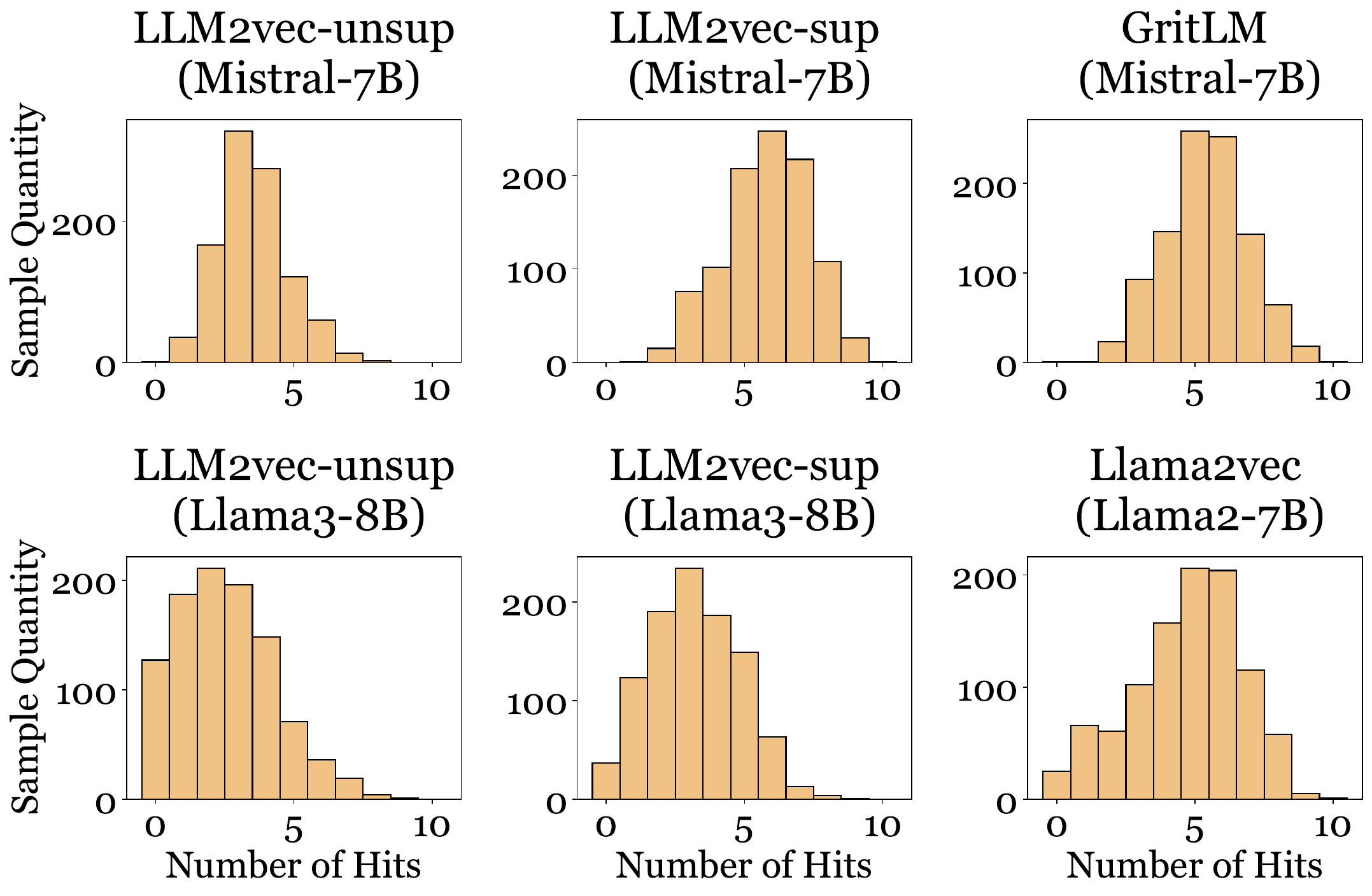}
    \caption{Per-Sentence Count of Top-10 Tokens Appearing}
    \label{fig:section4-sep}
\end{figure}
Figure \ref{fig:section4-all} presents the results of evaluating the top-10 tokens for each sample, yielding a total of 10,000 tokens (1,000 texts × 10 tokens). Tokens in the original text are labeled as Hit, while those absent are labeled as Miss. The two categories are roughly balanced, with a slightly higher count of tokens falling into the Miss category. Figure~\ref{fig:section4-sep} shows, for each text, how many of its top-10 tokens appear in the text itself. We can observe that in each text, there are 4 to 6 tokens that rank among the top 10.
The two statistical analyses suggest that the top 10 tokens are consistently present in both the tokens within the texts and those outside the texts.

\begin{table*}[htbp]
\centering
    \small
    \resizebox{0.9\textwidth}{!}{
    
    \begin{tabular}{@{} >{\centering\arraybackslash}p{0.10\linewidth} |
                    >{\centering\arraybackslash}p{0.15\linewidth} |
                    >{\centering\arraybackslash}p{0.75\linewidth} @{}}

    \toprule
    \textbf{BackBone} & \textbf{Model} & \textbf{Top 10 Aligned Tokens} \\
    \midrule
     &LLM2Vec-unsup &
      \tok{myello!70}{\_Ball}\;
      \tok{myello!70}{\_Gold}\;
      \tok{myred!70}{\_gold}\;
      \tok{myred!70}{Gold}\;
      \tok{myred!70}{gold}\;
      \tok{myred!70}{ball}\;
      \tok{myred!70}{\_ball}\;
      \tok{myello!70}{\_South}\;
      \tok{myred!70}{South}\;
      \tok{myred!70}{\_balls} \\

    Mistral-7B &LLM2Vec-sup &
      \tok{myello!70}{\_South}\;
      \tok{myello!70}{\_Gold}\;
      \tok{myello!70}{\_gold}\;
      \tok{myred!70}{\_migration}\;
      \tok{myred!70}{\_Ball}\;
      \tok{myello!70}{\_discovery}\;
      \tok{mygrey!70}{\_}\;
      \tok{myello!70}{\_migr}\;
      \tok{myred!70}{\_Australia}\;
      \tok{myred!70}{\_discover} \\

    &GritLM &
      \tok{myello!70}{\_South}\;
      \tok{myello!70}{\_Ball}\;
      \tok{myello!70}{\_Gold}\;
      \tok{myello!70}{\_migration}\;
      \tok{myred!70}{\_ball}\;
      \tok{mygrey!70}{\_}\;
      \tok{myred!70}{\_gold}\;
      \tok{myred!70}{\_SA}\;
      \tok{myred!70}{\_migr}\;
      \tok{myello!70}{2} \\
      \midrule
   
    \multirow{2}{*}{Lllama3-8B} &LLM2Vec-unsup &
      \tok{myred!70}{\_gold}\;
      \tok{myello!70}{\_Gold}\;
      \tok{myred!70}{ustr}\;
      \tok{myred!70}{\_Sou}\;
      \tok{myello!70}{\_had}\;
      \tok{myred!70}{\_sou}\;
      \tok{myello!70}{\_South}\;
      \tok{myred!70}{\_gol}\;
      \tok{myred!70}{\_golf}\;
      \tok{myred!70}{\_ever} \\
      
    &LLM2Vec-sup &
      \tok{myello!70}{\_Gold}\;
      \tok{myred!70}{\_gold}\;
      \tok{myred!70}{\_Discovery}\;
      \tok{myello!70}{\_discovery}\;
      \tok{myred!70}{Gold}\;
      \tok{myred!70}{gold}\;
      \tok{myred!70}{\_Discover}\;
      \tok{myred!70}{\_discover}\;
      \tok{myello!70}{\_had}\; \tok{myred!70}{\_discovered}\ \\
    \midrule
    Llama2-7B &Llama2vec &
      \tok{myred!70}{\_gold}\;
      \tok{myello!70}{\_Gold}\;
      \tok{mygrey!70}{?}\;
      \tok{mygrey!70}{\_}\;
      \tok{myello!70}{\_Australia}\;
      \tok{mygrey!70}{.}\;
      \tok{mygrey!70}{”.}\;
      \tok{myello!70}{\_South}\;
      \tok{mygrey!70}{!}\;
      \tok{mygrey!70}{\_’} \\
    \midrule
     \multicolumn{3}{p{0.90\linewidth}}{%
      Text: The discovery of Gold in Ballarat caused a large migration from South Australia and by 1852 some 8000 had left for the Goldfields.
    } \\
    \bottomrule
  \end{tabular}
  }
  
  \centering
\end{table*}

\begin{table*}[htbp]
\centering
    \small
    \resizebox{0.9\textwidth}{!}{
    
    \begin{tabular}{@{} >{\centering\arraybackslash}p{0.10\linewidth} |
                    >{\centering\arraybackslash}p{0.15\linewidth} |
                    >{\centering\arraybackslash}p{0.75\linewidth} @{}}

    \toprule
    \textbf{BackBone} & \textbf{Model} & \textbf{Top 10 Aligned Tokens} \\
    \midrule
     &LLM2Vec-unsup &
      \tok{myello!70}{\_Work}\;
      \tok{myred!70}{Work}\;
      \tok{myello!70}{\_Allen}\;
      \tok{myred!70}{\_aller}\;
      \tok{myello!70}{\_Services}\;
      \tok{myred!70}{\_Works}\;
      \tok{myred!70}{WORK}\;
      \tok{myred!70}{allen}\;
      \tok{myred!70}{\_allerg}\;
      \tok{myred!70}{Worker} \\

    Mistral-7B &LLM2Vec-sup &
      \tok{myello!70}{\_Work}\;
      \tok{myello!70}{\_Man}\;
      \tok{myello!70}{\_work}\;
      \tok{myred!70}{Work}\;
      \tok{myred!70}{\_Australian}\;
      \tok{myello!70}{\_man}\;
      \tok{myred!70}{\_cricket}\;
      \tok{myello!70}{\_War}\;
      \tok{myred!70}{\_cr}\;
      \tok{myred!70}{\_Australia} \\

    &GritLM &
      \tok{myred!70}{\_Work}\;
      \tok{myello!70}{\_Services}\;
      \tok{myred!70}{\_Australian}\;
      \tok{myello!70}{Work}\;
      \tok{myred!70}{\_cricket}\;
      \tok{myred!70}{\_services}\;
      \tok{myello!70}{\_cr}\;
      \tok{myello!70}{\_James}\;
      \tok{myello!70}{\_Jim}\;
      \tok{myred!70}{\_work} \\
      \midrule

    \multirow{2}{*}{Lllama3-8B} &LLM2Vec-unsup &
      \tok{myello!70}{\_cricket}\;
      \tok{myred!70}{a}\;
      \tok{myred!70}{XHR}\;
      \tok{mygrey!70}{G}\;
      \tok{mygrey!70}{ustr}\;
      \tok{mygrey!70}{icit}\;
      \tok{mygrey!70}{eper}\;
      \tok{myred!70}{ohan}\;
      \tok{myred!70}{ellen}\;
      \tok{myred!70}{\_pu} \\
  
    &LLM2Vec-sup &
      \tok{myello!70}{\_cricket}\;
      \tok{myred!70}{\_Cricket}\;
      \tok{myello!70}{\_Work}\;
      \tok{myred!70}{\_work}\;
      \tok{myred!70}{\_Australia}\;
      \tok{myello!70}{\_cr}\;
      \tok{myred!70}{GA}\;
      \tok{myred!70}{\_Cr}\;
      \tok{myello!70}{\_}\;
      \tok{myello!70}{\_Australian}\ \\
    \midrule
    Llama2-7B &Llama2vec &
      \tok{myello!70}{\_Work}\;
      \tok{myello!70}{\_Australian}\;
      \tok{myello!70}{\_cr}\;
      \tok{myello!70}{\_James}\;
      \tok{myello!70}{\_Services}\;
      \tok{mygrey!70}{\_}\;
      \tok{myred!70}{\_work}\;
      \tok{myred!70}{\_Cr}\;
      \tok{myred!70}{cr}\;
      \tok{mygrey!70}{?} \\
    \midrule
     \multicolumn{3}{p{0.90\linewidth}}{%
      Text: James Allen Workman (17 March 1917 – 23 December 1970) was an Australian cricketer who played first-class cricket for the Australian Services team from May 1945 to January 1946.
    } \\
    \bottomrule
  \end{tabular}
  }
  \centering
\end{table*}

\begin{table*}[htbp]
\centering
    \small
    \resizebox{0.9\textwidth}{!}{
    
    \begin{tabular}{@{} >{\centering\arraybackslash}p{0.10\linewidth} |
                    >{\centering\arraybackslash}p{0.15\linewidth} |
                    >{\centering\arraybackslash}p{0.75\linewidth} @{}}

    \toprule
    \textbf{BackBone} & \textbf{Model} & \textbf{Top 10 Aligned Tokens} \\
    \midrule
     &LLM2Vec-unsup &
      \tok{myello!70}{\_Cow}\;
      \tok{myred!70}{\_cow}\;
      \tok{myello!70}{\_Henry}\;
      \tok{myred!70}{cow}\;
      \tok{myred!70}{Hen}\;
      \tok{myred!70}{\_hen}\;
      \tok{myred!70}{\_Hen}\;
      \tok{myred!70}{\_punk}\;
      \tok{myred!70}{\_Henri}\;
      \tok{myred!70}{\_revolutionary} \\
  
    Mistral-7B &LLM2Vec-sup &
      \tok{myred!70}{\_problem}\;
      \tok{myello!70}{\_Cow}\;
      \tok{myello!70}{\_Henry}\;
      \tok{myred!70}{\_world}\;
      \tok{myred!70}{\_cow}\;
      \tok{myello!70}{\_World}\;
      \tok{myello!70}{\_music}\;
      \tok{myello!70}{\_Problem}\;
      \tok{myred!70}{\_problems}\;
      \tok{myred!70}{\_} \\
  
    &GritLM &
      \tok{myred!70}{\_Henry}\;
      \tok{myello!70}{\_problem}\;
      \tok{myred!70}{\_world}\;
      \tok{myello!70}{\_Cow}\;
      \tok{myred!70}{\_cow}\;
      \tok{myred!70}{\_World}\;
      \tok{myello!70}{\_music}\;
      \tok{myello!70}{Hen}\;
      \tok{myello!70}{\_Problem}\;
      \tok{myred!70}{\_history} \\
      \midrule
  
    \multirow{2}{*}{Lllama3-8B} &LLM2Vec-unsup &
      \tok{myred!70}{\_problems}\;
      \tok{myred!70}{\_problem}\;
      \tok{myello!70}{\_Cow}\;
      \tok{myello!70}{\_Henry}\;
      \tok{myred!70}{\_cow}\;
      \tok{myello!70}{\_Problem}\;
      \tok{myred!70}{\_Problems}\;
      \tok{myred!70}{\_cows}\;
      \tok{myred!70}{\_world}\;
      \tok{myred!70}{\_problematic} \\
  
    &LLM2Vec-sup &
      \tok{myred!70}{\_problem}\;
      \tok{myred!70}{\_Cow}\;
      \tok{myello!70}{\_Henry}\;
      \tok{myello!70}{\_Problem}\;
      \tok{myred!70}{\_problems}\;
      \tok{myred!70}{\_cow}\;
      \tok{myello!70}{\_world}\;
      \tok{myred!70}{\_World}\;
      \tok{myred!70}{\_Problems}\;
      \tok{myred!70}{Cow}\ \\
    \midrule
    Llama2-7B &Llama2vec &
      \tok{myello!70}{\_Henry}\;
      \tok{myello!70}{\_Cow}\;
      \tok{mygrey!70}{\_}\;
      \tok{myred!70}{\_cow}\;
      \tok{mygrey!70}{?}\;
      \tok{myred!70}{\_hen}\;
      \tok{mygrey!70}{.}\;
      \tok{myred!70}{cow}\;
      \tok{myred!70}{\_world}\;
      \tok{mygrey!70}{!} \\
    \midrule
     \multicolumn{3}{p{0.90\linewidth}}{%
      Text: "The World Is a Problem" chronicles the history of Henry Cow and their exploration of music and activism, from their inception in 1968 to their break-up in 1978.
    } \\
    \bottomrule
  \end{tabular}
  }
  \centering
    \caption{Top-10 Aligned Token from the Text2Token Framework on different embedders.}
     \label{tab:section4-sota_plus}
\end{table*}

\begin{table*}[htbp]
\centering
    \small
    \resizebox{0.9\textwidth}{!}{
    
    \begin{tabular}{@{} >{\centering\arraybackslash}m{0.2\linewidth} |     >{\centering\arraybackslash}m{0.75\linewidth} @{}} 
    \toprule
    \textbf{Model} & \textbf{Top 10 Aligned Tokens} \\
    \midrule
    Raw (Mistral-7B)&
      \tok{mygrey!70}{\textbackslash n}\;\tok{mygrey!70}{\_n}\;\tok{mygrey!70}{,}\;\tok{myello!70}{\_Gold}\;\tok{myello!70}{\_Australia}\;\tok{mygrey!70}{.}\;\tok{myello!70}{\_in}\;\tok{myello!70}{\_and}\;\tok{myred!70}{\_to}\;\tok{myello!70}{1} \\
  
    Text2Token (Mistral-7B) &
      \tok{myello!70}{\_Ball}\;\tok{myello!70}{\_Gold}\;\tok{myred!70}{\_gold}\;\tok{myred!70}{\_Victoria}\;\tok{myred!70}{\_Victorian}\;\tok{myred!70}{\_Melbourne}\;\tok{myred!70}{\_ball}\;\tok{myello!70}{\_Australia}\;\tok{myred!70}{Gold}\;\tok{myred!70}{\_Australian} \\

    \midrule
    Raw (Llama3-8B) &
      \tok{mygrey!70}{,}\;\tok{myello!70}{\_the}\;\tok{myello!70}{\_}\;\tok{mygrey!70}{.}\;\tok{myello!70}{\_Ball}\;\tok{mygrey!70}{$C$}\;\tok{mygrey!70}{$...C$}\;\tok{myello!70}{\_of}\;\tok{myello!70}{\_in};\tok{mygrey!70}{$A$}\ \\
    Text2Token (Llama3-8B) &
      \tok{myred!70}{Victoria}\;\tok{myello!70}{\_Ball}\;\tok{myred!70}{Ball}\;\tok{myred!70}{Gold}\;\tok{myred!70}{\_Victoria}\;\tok{myred!70}{Australia}\;\tok{myred!70}{gold}\;\tok{myello!70}{185}\;\tok{myred!70}{\_miners}\;\tok{myred!70}{\_dig} \\
    \midrule
     \multicolumn{2}{p{0.95\linewidth}}{%
      Text: The discovery of Gold in Ballarat caused a large migration from South Australia and by 1852 some 8000 had left for the Goldfields.
    } \\
    \bottomrule
  \end{tabular}
  }
\end{table*}
\begin{table*}[htbp]
\centering
    \small
    \resizebox{0.9\textwidth}{!}{

    \begin{tabular}{@{} >{\centering\arraybackslash}m{0.2\linewidth} |     >{\centering\arraybackslash}m{0.75\linewidth} @{}} 
    \toprule
    \textbf{Model} & \textbf{Top 10 Aligned Tokens} \\
    \midrule
    Raw (Mistral-7B)&
      \tok{myello!70}{1}\;\tok{mygrey!70}{\_}\;\tok{mygrey!70}{\_n}\;\tok{mygrey!70}{,}\;\tok{myred!70}{.}\;\tok{myello!70}{2}\;\tok{myred!70}{\_and}\;\tok{mygrey!70}{\_(}\;\tok{mygrey!70}{\_in}\;\tok{mygrey!70}{...} \\
  
    Text2Token (Mistral-7B) &
      \tok{myello!70}{\_Work}\;\tok{myello!70}{\_cricket}\;\tok{myello!70}{\_Allen}\;\tok{myello!70}{\_Services}\;\tok{myred!70}{Work}\;\tok{myred!70}{\_Works}\;\tok{myello!70}{\_Australian}\;\tok{myred!70}{\_Cr}\;\tok{myello!70}{\_James}\;\tok{myred!70}{\_Test} \\

    \midrule
    Raw (Llama3-8B) &
      \tok{mygrey!70}{\_}\;\tok{mygrey!70}{...C}\;\tok{mygrey!70}{...}\;\tok{mygrey!70}{\_...}\;\tok{mygrey!70}{,}\;\tok{mygrey!70}{...C}\;\tok{mygrey!70}{C}\;\tok{mygrey!70}{.}\;\tok{mygrey!70}{\_and};\tok{mygrey!70}{...CC}\ \\
    Text2Token (Llama3-8B) &
      \tok{myello!70}{\_cricket}\;\tok{myred!70}{Work}\;\tok{myello!70}{\_Work}\;\tok{myello!70}{\_cr}\;\tok{myred!70}{cr}\;\tok{myred!70}{Australian}\;\tok{myred!70}{Allen}\;\tok{myred!70}{\_Cricket}\;\tok{myred!70}{Australia}\;\tok{myello!70}{\_Australian} \\
    \midrule
     \multicolumn{2}{p{0.95\linewidth}}{%
      Text: James Allen Workman (17 March 1917 – 23 December 1970) was an Australian cricketer who played first-class cricket for the Australian Services team from May 1945 to January 1946.
    } \\
    \bottomrule
  \end{tabular}
  }
\end{table*}
\begin{table*}[htbp]
\centering
    \small
    \resizebox{0.9\textwidth}{!}{
    
    \begin{tabular}{@{} >{\centering\arraybackslash}m{0.2\linewidth} |     
    >{\centering\arraybackslash}m{0.75\linewidth} @{}} 
    \toprule
    \textbf{Model} & \textbf{Top 10 Aligned Tokens} \\
    \midrule
    Raw (Mistral-7B)&
      \tok{mygrey!70}{,}\;\tok{myred!70}{\_the}\;\tok{mygrey!70}{\_}\;\tok{mygrey!70}{\_"}\;\tok{myello!70}{\_in}\;\tok{myred!70}{\_cow}\;\tok{mygrey!70}{\textbackslash n}\;\tok{mygrey!70}{\_a}\;\tok{mygrey!70}{.}\;\tok{myello!70}{\_Problem} \\
  
    Text2Token (Mistral-7B) &
      \tok{myello!70}{\_Cow}\;\tok{myello!70}{\_Henry}\;\tok{myred!70}{Hen}\;\tok{myred!70}{\_cow}\;\tok{myred!70}{\_Hen}\;\tok{myred!70}{cow}\;\tok{myello!70}{\_Problem}\;\tok{myred!70}{\_problem}\;\tok{myred!70}{\_Crow}\;\tok{myred!70}{\_hen} \\

    \midrule
    Raw (Llama3-8B) &
      \tok{mygrey!70}{...}\;\tok{myred!70}{\_and}\;\tok{mygrey!70}{\_the}\;\tok{mygrey!70}{.}\;\tok{myello!70}{,}\;\tok{myred!70}{\_cow}\;\tok{mygrey!70}{\_...H}\;\tok{myello!70}{\_Henry}\;\tok{mygrey!70}{K}\;\tok{myello!70}{\_in} \\
    Text2Token (Llama3-8B) &
      \tok{myello!70}{\_Cow}\;\tok{myred!70}{Cow}\;\tok{myred!70}{Henry}\;\tok{myello!70}{\_Henry}\;\tok{myred!70}{punk}\;\tok{myred!70}{\_cow}\;\tok{myred!70}{\_bands}\;\tok{myred!70}{music}\;\tok{myred!70}{\_punk}\;\tok{myred!70}{cow} \\
       \midrule
     \multicolumn{2}{p{0.95\linewidth}}{
      Text: "The World Is a Problem" chronicles the history of Henry Cow and their exploration of music and activism, from their inception in 1968 to their break-up in 1978.
    } \\
    \bottomrule
  \end{tabular}
  }
  \centering
  \caption{Top-10 Aligned Token from the Text2Token Framework on raw backbone and our embedders for Three Examples.}
  \label{tab:our_more_examples}
\end{table*}
\section{Filter Formula Comparison}
To filter out noisy elements in the raw token distribution from the untuned model, we evaluate three filtering formulas. The effect is illustrated by a sample in Table~\ref{tab:filter_formula}. We ultimately adopted the Div type for the main study.
\label{sec:filter_comparsion}

\begin{table*}[htbp]
\centering
    \small
    \resizebox{0.9\textwidth}{!}{
    
    \begin{tabular}{@{} >{\centering\arraybackslash}m{0.1\linewidth} |
    >{\centering\arraybackslash}m{0.3\linewidth} | 
    >{\centering\arraybackslash}m{0.55\linewidth} @{}} 
    
    \toprule
    \textbf{Model} & \textbf{Filter Formula} &\textbf{Top 10 Aligned Tokens} \\
    \midrule
    Original  & - & 
      \tok{mygrey!70}{.}\;
      \tok{mygrey!70}{\_the}\;
      \tok{mygrey!70}{. \textbackslash n}\;
      \tok{mygrey!70}{\_a}\;
      \tok{mygrey!70}{,}\;
      \tok{mygrey!70}{\_in}\;
      \tok{mygrey!70}{\_and}\;
      \tok{mygrey!70}{...}\;
      \tok{mygrey!70}{... \textbackslash n}\;
      \tok{myello!70}{\_file} \\ \midrule
    Log type  & $\hat{y}_{w_k} = - Q_{\theta_0}(w_k | x_i) \log \overline{Q}_{\theta_0}(w_k)$ & 
      \tok{mygrey!70}{.}\;
      \tok{mygrey!70}{\_a}\;
      \tok{mygrey!70}{. \textbackslash n}\;
      \tok{mygrey!70}{\_the}\;
      \tok{myello!70}{\_file}\;
      \tok{myello!70}{\_corruption}\;
      \tok{myred!70}{\_data}\;
      \tok{mygrey!70}{\_may}
      \tok{mygrey!70}{\_in}\;
      \tok{mygrey!70}{...}\; \\ \midrule
    Div type  &  $\hat{y}_{w_k} = \frac{Q_{\theta_0}(w_k | x_i)}{Q_{\theta_0}(w_k | x_i) + \overline{Q}_{\theta_0}(w_k)}$ & 
      \tok{myello!70}{\_corruption}\;
      \tok{myello!70}{\_Excel}\;
      \tok{myred!70}{\_corrupted}\;
      \tok{myred!70}{\_excel}\;
      \tok{myred!70}{\_corrupt}\;
      \tok{myred!70}{Corruption}\;
      \tok{myello!70}{\_shutdown}\;
      \tok{myello!70}{\_Ms}\;
      \tok{myred!70}{Microsoft}\;
      \tok{myello!70}{\_file} \\ \midrule
    Sub type  & $\hat{y}_{w_k} = \log Q_{\theta_0}(w_k | x_i) - \log \overline{Q}_{\theta_0}(w_k)$ & 
      \tok{mygrey!70}{.}\;
      \tok{mygrey!70}{\_a}\;
      \tok{mygrey!70}{. \textbackslash n}\;
      \tok{mygrey!70}{the}\;
      \tok{myello!70}{\_file}\;
      \tok{myred!70}{\_data}\;
      \tok{mygrey!70}{may}\;
      \tok{myello!70}{\_corruption}\;
      \tok{mygrey!70}{...}\;
      \tok{myello!70}{\_can} \\ \midrule
    \multicolumn{3}{p{0.95\linewidth}}{%
      Text: Common reasons for Ms excel corruption. Improper shutdown of computer a Basically the files size of Excel are large therefore if there is any improper shutdown of system occur there are chances that your open Ms excel can easily corrupt. The improper shut down can occur due to power failure or any other reason.
    } \\
    \bottomrule
  \end{tabular}
  }
  \centering
    \caption{Example of three filtering formulas on the Top-10 Aligned Tokens in the Text2Token framework.}

    \label{tab:filter_formula}
\end{table*}

\begin{table*}[ht]
    \centering
    \small
     \resizebox{0.9\textwidth}{!}{
    \begin{tabular}{lll|lll}
    \toprule
        Task & Mistral-7B & Llama3-8B & Task & Mistral-7B & Llama3-8B \\ 
        \midrule
        Banking77Classification & 82.37 & 81.73 & ArguAna & 52.73 & 55.03 \\ 
        ImdbClassification & 67.40 & 67.32 & CQADupstackGamingRetrieval & 39.22 & 33.53 \\ 
        MTOPDomainClassification & 92.72 & 93.26 & CQADupstackUnixRetrieval & 20.91 & 27.56 \\
        MassiveIntentClassification & 69.32 & 69.15 & ClimateFEVERHardNegatives & 23.53 & 28.48 \\ 
        MassiveScenarioClassification & 75.63 & 76.10 & FEVERHardNegatives & 62.81 & 71.40 \\ 
        ToxicConversationsClassification & 63.87 & 65.75 & FiQA2018 & 29.70 & 28.62 \\
        TweetSentimentExtractionClassification & 54.59 & 52.39 & HotpotQAHardNegatives & 49.49 & 51.37 \\ 
        AmazonCounterfactualClassification & 66.16 & 62.18 & SCIDOCS & 18.12 & 18.56 \\
        ArXivHierarchicalClusteringP2P & 61.12 & 62.47 & TRECCOVID & 61.55 & 57.10 \\ 
        ArXivHierarchicalClusteringS2S & 59.98 & 61.99 & Touche2020Retrieval.v3 & 40.57 & 44.03 \\ 
        BiorxivClusteringP2P.v2 & 42.20 & 42.74 & BIOSSES & 81.98 & 82.75 \\ 
        MedrxivClusteringP2P.v2 & 36.13 & 37.18 & SICK-R & 69.19 & 68.32 \\
        MedrxivClusteringS2S.v2 & 33.79 & 36.93 & STS12 & 56.13 & 64.24 \\ 
        StackExchangeClustering.v2 & 66.63 & 68.57 & STS13 & 72.45 & 78.14 \\ 
        StackExchangeClusteringP2P.v2 & 44.01 & 47.95 & STS14 & 67.80 & 71.11 \\
        TwentyNewsgroupsClustering.v2 & 44.64 & 54.22 & STS15 & 78.45 & 79.64 \\ 
        SprintDuplicateQuestions & 89.26 & 91.69 & STSBenchmark & 73.83 & 74.34 \\ 
        TwitterSemEval2015 & 58.21 & 55.3 & STS17 & 82.50 & 80.91 \\ 
        TwitterURLCorpus & 79.36 & 81.29 & STS22.v2 & 64.66 & 56.32 \\ 
        AskUbuntuDupQuestions & 55.75 & 57.44 & SummEvalSummarization.v2 & 10.05 & 28.16 \\ 
        MindSmallReranking & 33.39 & 33.76 & Average & 55.10 & 55.25 \\ \bottomrule
    \end{tabular}
    }
\caption{Details results of Text2Token on MTEB v2.}
\label{main_detail}
\end{table*}
\section{Aligned Token Analysis}
\label{Aligned Token}
To analyze the ability of aligned tokens in our model, we provide three more examples in Table \ref{tab:our_more_examples} and observe similar phenomena as in Section~\ref{sec:analysis}.

\section{Detailed Results}
\label{Detailed Results}
Table~\ref{main_detail} presents the detailed results of the two models that performed the best in the main results.

\end{document}